\pdfoutput=1
\documentclass[11pt]{article}

\newcommand{\revoption}{final}
\usepackage[\revoption]{acl}

\usepackage{times}
\usepackage{latexsym}

\usepackage{float}
\usepackage{graphicx}
\usepackage{subcaption}
\usepackage{xcolor}
\usepackage{hyperref}
\usepackage{xcolor}		% make links dark blue
 \definecolor{darkblue}{rgb}{0, 0, 0.5}
 \hypersetup{colorlinks=true,citecolor=darkblue, linkcolor=darkblue, urlcolor=darkblue}

\def\Snospace~{\S{}} %makes section autorefs more beautiful

\usepackage[T1]{fontenc}
\usepackage{microtype}

\usepackage{adjustbox}
\usepackage{graphicx}
\usepackage{multirow}
\usepackage{booktabs}
\usepackage{todonotes}
\usepackage{float}
\usepackage{xstring}
\usepackage{enumitem}
\usepackage[bottom]{footmisc}

\definecolor{lightblue}{HTML}{E0ECF7}
\definecolor{darkblue}{HTML}{092E6B}
\usepackage{adjustbox}
\usepackage{array}
\usepackage{booktabs}
\usepackage{multirow}
\usepackage{amssymb}
\newcolumntype{R}[2]{%
    >{\adjustbox{angle=#1,lap=\width-(#2)}\bgroup}%
    l%
    <{\egroup}%
}
\title{The HCI Aspects of Public Deployment of Research Chatbots:\\ A User Study, Design Recommendations, and Open Challenges}

\author{\normalfont Morteza Behrooz, \\William Ngan, Joshua Lane, Giuliano Morse, Benjamin Babcock, Kurt Shuster, \\Mojtaba Komeili, Moya Chen, Melanie Kambadur, Y-Lan Boureau, Jason Weston \\ Meta AI}

\begin{document}
\maketitle
\begin{abstract}
Publicly deploying research chatbots is a nuanced topic involving necessary risk-benefit analyses. While there have recently been frequent discussions on whether it is responsible to deploy such models, there has been far less focus on the interaction paradigms and design approaches that the resulting interfaces should adopt, in order to achieve their goals more effectively. We aim to pose, ground, and attempt to answer HCI questions involved in this scope, by reporting on a mixed-methods user study conducted on a recent research chatbot. We find that abstract anthropomorphic representation for the agent has a significant effect on user's perception, that offering AI explainability may have an impact on feedback rates, and that two (\textit{diegetic} and \textit{extradiegetic}) levels of the chat experience should be intentionally designed. We offer design recommendations and areas of further focus for the research community.
\end{abstract}

\section{Introduction}

Given the recent advances in generative Large Language Models (LLMs), publicly available modern chatbots are becoming more commonplace and popular \cite{chatgptPerceptions, bb3, lamda}. Releasing the underlying models, or openly deploying such chatbots for the general public is a complicated decision involving necessary and nuanced considerations in terms of safety and responsible research conduct. An argument that may have led to the recent surge in deployments focuses on the need to acquire diverse and wide-ranging user feedback data in an isolated and research-only setting, in order to later improve the model's safety and capability. Moreover, evaluating chatbots becomes increasingly important as they rapidly improve, while scoped, limited, and ``clinical'' evaluation of them is proving unreliable \cite{chatbotEvalOpenProblem, evaluationmethods}. As an alternative or complementing approach, ecologically valid user feedback originated from organic in-situ experience can establish more suitable evaluations. Such an approach may enable a better estimation of the model capabilities, more rigorously uncover limitations such as safety risks, shine a light on ways in which users may bypass safety mechanisms \cite{chatgptsafetybypass}, show unintended societal or domain-specific consequences \cite{chatgptExams}, point to suitable use cases \cite{chatgptPerceptions}, etc. 

Without entering a discussion of the arguments for or against the public release of research chatbots, in this work, we recognize that such chatbots are now both widely available and increasingly popular \cite{chatgptPerceptions, bb3, lamda}. We further recognize that the interaction paradigms and design decisions used in the deployment of such chatbots are very likely to affect many aspects of the user's experience: their perception of the chat and of the agent, their enjoyment levels, chat topics, chat lengths, user retention, propensity to give feedback, public perception of what ``AI'' is, and generally, the societal narrative around technology, among others. We propose that the interface and interaction design decisions are 1) non-trivial, and 2) consequential.

\subsection{Motivation}

We define the group of chatbots in focus as advanced, LLM-based, textual conversational agents (CAs) that are released to the public as an interactive experience, such OpenAI's ChatGPT, Meta AI's BlenderBot, and Google's LaMDA. Crucially, there are a set of research goals in releasing such chatbots, and while they vary, the set usually includes: 

\begin{itemize}
    \item Showcasing a new model's capability to impress, or increase research transparency
    \item Acquiring conversational data (i.e. chat logs) to improve the model
    \item Acquiring user feedback data (e.g. thumbs up/down) to later improve the model
    \item Finding points of safety failure, inappropriate or favorable usage, future use cases, and other scale-dependent discoveries
    \item Enabling better experimental setups, such as A/B testing, to help evaluate models
\end{itemize}

Besides the generic impacts of design on user's experience and behavior, it is important to recognize that design can directly impact the above goals. For instance, research transparency and democratization of AI would both have to involve proper levels of education and explainability to be successful. Design is also likely to affect user's propensity to provide feedback, a frequent and significant item in deployments' rationales. As a parallel, the literature in Intelligent Virtual Agents (IVAs) and Human-Robot Interaction (HRI) communities underline both the complexities and consequences of the design decisions in their respective areas too \cite{iva}.

 We thus ask: \textit{What are the HCI questions, answers, and open challenges involved in deploying research chatbots to the public?} It's noteworthy that the involved design questions would persist if a research chatbot was released to a limited population or a specific group of users, even if the answers vary for some very specific populations or use cases. In any case, changing the release model of conversational agents would not remove the need to ask and solve the inherent HCI challenges. In the following sections, we study how users would interact with research-focused chatbots, list the likely cognitive factors involved, and attempt to understand what design approaches may contribute to solving the resulting issues. 

\subsection{Root Cognitive Factors}
\label{cogfactors}

Given that certain fixed factors exist in deploying research chatbots, they create a fixed set of root challenges in need of design solutions that are informed and intentional.

\subsubsection{Lack of Conversational Context}
\label{lackcontext}

What is consistent and inherent in research chatbot deployments is \textit{a lack of user journey}. While there does exist an implicit context of ``experimentation with a new AI'', this does not provide for much more conversational or cognitive grounding than an initial novelty effect or curiosity. Human conversations occur in contexts, and specifically, cognitive processes engaged in conversation are significantly impacted by the context \cite{conversationcontext}. While conversational context is a dynamic process that can be partial, relative, and subjective, it does cognitively command the communicative process and experience for human conversants \cite{commentsoncontext}. 

The next two factors collectively point to a lack of mental model in the human user about who the chatbot agent is, and what it may be cognitively and conversationally capable of. We focus on the cognitive impact of this lack of mental model, as it enables a more grounded outlook and set of hypotheses to address.

\subsubsection{The ``Speaker Perception Void''}
\label{speakervoid}

Humans necessarily develop a perception of the parties with whom they are conversing. This reality is rooted in human cognition. In \cite{speakerperception}, an attributional model of conversational inference is presented that shows how listener's message interpretations are guided by their perceptions of the speaker. Further grounding can be found in cognitive psychology, where the theory of social cognition is shown to be intertwined with language use when it comes to interaction \cite{socialcog}. It is plausible to assume that human-agent textual conversations will inherit much of the same effects, given the constant nature of human cognition. We can then argue that the perceptions humans develop from chatbots are almost entirely dependent on conversations and interface (bar the generic ``AI'' perception that may bring a priori perceptions). Thus, the burden of anchoring the user's inevitable perception of the chatbot is at least partly on the interface, interaction, and experience design, and this user perception is likely to affect conversational content, engagingness, enjoyment level, feedback rates, chat lengths, and other UX or deployment goals.

\subsubsection{Lack of an Expectation Baseline}
\label{expectationbaseline}

This effect is closely related to the speaker perception void, and perhaps a specific yet significant aspect of it, causing an inability in the human user to reliably apply her natural assumptions about the communicative and cognitive capabilities of a human conversational partner onto a chatbot. Similar to how language capabilities are hard to communicate in voice assistants, potentially causing an interface exploration issue \cite{voiceUIissues}, research chatbots do not reliably communicate their communicative and cognitive capabilities, leaving the human users with two distinct and simultaneous tasks: conversing and evolving a picture of how they can converse. This results in cognitive load and may negatively impact the experience, if unguided. While humans as conversational partners do vary in their communicative style, they usually meet a baseline of shared expectations. This shared understanding is so innate that it contributes to causing \textit{socio-cognitive development} in children \cite{kidsTOMchat}. However, chatbots frequently break these shared, innate expectations through deviation and error: making stylistic or factual mistakes, unnatural responses and transitions, nonsensical statements, and more. This effect comes close to causing ``conversational uncanny valley'' where users may perceive the chat to be strange, or explicitly ``uncanny'', especially in peripheral interaction and interface elements such as response wait times \cite{chatbotuncannyvalley}. A lack of cognitive expectation is not unique to research chatbots, yet it especially applies to them because it is confounded by a lack of context and a general lack of speaker perception. We believe that here, too, interaction and interface design can help communicate expectations, recover from deviations \cite{chatboterrors, amimeoryou} and augment the user's conversational experience such that the goals of research chatbots are better achieved. 

\subsection{Other Related Work}

Much of the previous work focuses on the chatbots' conversational UX in terms of the chat qualities itself. For instance, in \cite{chatbotTypology}, a typology of utility-oriented chatbots is offered along with recommendations on how the conversation and chat should occur in various types. While conversational UX (i.e. designing what is said in messages) is a main driver of user experience in chatbots. A few important vectors of conversational UX are frequently studied, such as the level of sociability and utility \cite{chatbotsociabilityandutility}, the effects of having memory in chat \cite{xu2021beyond}, or internet search \cite{chatbotsearch}, among others.

While it has attracted less attention than the conversational UX \cite{folstad2017chatbots}, there is previous work focused on other HCI aspects of chatbot interaction, such as the interface design, agent representation, non-textual conversational inputs \cite{emoji1,emoji2}, and other elements. This is especially relevant to research chatbots given the cognitive factors above. 

Visual chatbot representation can take many forms, from physical and robotic, to humanlike virtual agents \cite{iva}, and many forms of other static or animated representations that can accompany a mainly textual chat experience. A significant aspect of anthropomorphism would depend on the chat style itself \cite{anthropomorphic1, anthropomorphic2}. However, representation of the agent carries a large effect too; for instance, Replika, as a popular commercialized chatbot uses 3D humanlike avatars that show facial expressions \cite{replika}. Jung et al. use the Great Chain of Being framework to investigate the impact of using various non-human metaphors to represent chatbots \cite{metaphoricalrep}. 

One of the challenges in agent representation in research chatbots is how the specificity of representation may bias chats, and hence the resulting dataset, in many ways, e,g. towards properties semantically related to that representation. For instance, if the chatbot is a frog, it is very likely that many conversations (which happen outside other contexts and user flows) will relate to topics relevant to frogs or encode other attitudinal biases in them. Very elaborate humanlike representations (e.g. a virtual agent) may result in specific character implications, and bias tone, topic, and other aspects of chats as well. In \cite{behrooz2018cognitive}, authors experiment with the perception of interest in stories that involve abstract shapes. It would be interesting to study the impact of abstract and subtle anthropomorphic cues for chatbot agents.

In \cite{chatbotMessenger}, while UI-related findings are limited due to using an existing medium (Facebook Messenger) for chatbot interaction, authors report their design recommendations such as offering users pre-written response options to reduce typing and a ``persistent display of certain handy information such as description of the chatbot capabilities''. The latter points to the lack of cognitive expectation mentioned earlier.  

Some related work has focused on the design and experience of chatbots in specific domains. In \cite{botforhealth}, a literature review of important aspects of chatbot interface design in the health domain is discussed. Our focus, in studying the interface design for research chatbots is similarly scoped for a specific \textit{category} of chatbot, but different in that the conversations are often open-ended and not about one specific \textit{topic} or domain.

\section{User Study}
\label{sec:study}

In order to understand the HCI challenges and best design practices for research chatbots, we conducted a mixed-methods study with a set of research questions informed by the motivations and grounding laid out earlier.

\begin{itemize}
    \item \textbf{RQ1. } What are the effects of the relevant cognitive factors inherent to the deployment of research chatbots discussed in Sec. \ref{cogfactors}? Are there design and HCI solutions for possible undesired effects?
    \item \textbf{RQ2. } What is the role of eXplainable AI (XAI) in research chatbot deployments? Does offering it improve the experience? If so, how should it be designed?
    \item \textbf{RQ3. } How can we benefit from the proven positive effects of anthropomorphic design in agent representation, without biasing the conversation and user perception?
\end{itemize}

In addition to the above, a secondary goal of this study was to perform a qualitative evaluation of the conversational quality of the model itself, not to directly assess the model's quality, but in order to understand the viability of the triangulated results, and finding \textit{orientations} for the quantitative findings (i.e. find reasons, expansions, clarifications) often asked at scale on crowd worker platforms such as Amazon's Mechanical Turk \cite{chatbotEvalOpenProblem}.

\subsection{Chatbot}
\label{sec:chatbot}

We used a recent LLM-based chatbot \textit{bb3deploy} in order to investigate the research questions above. The underlying model is capable of social conversations, searching the internet to incorporate findings inside the chat, and holding memories from the conversations to also use in the subsequent exchanges and conversations. Figure \ref{fig:chatscreen} below shows the main chat screen and various interface elements. ``Extradiegetic'' chat refers to the communicative text inside the chat UI that is coming from the interface or system, and not the chatbot.

\begin{figure}
     \centering
     \begin{subfigure}{0.2\textwidth}
         \centering
         \includegraphics[width=\textwidth]{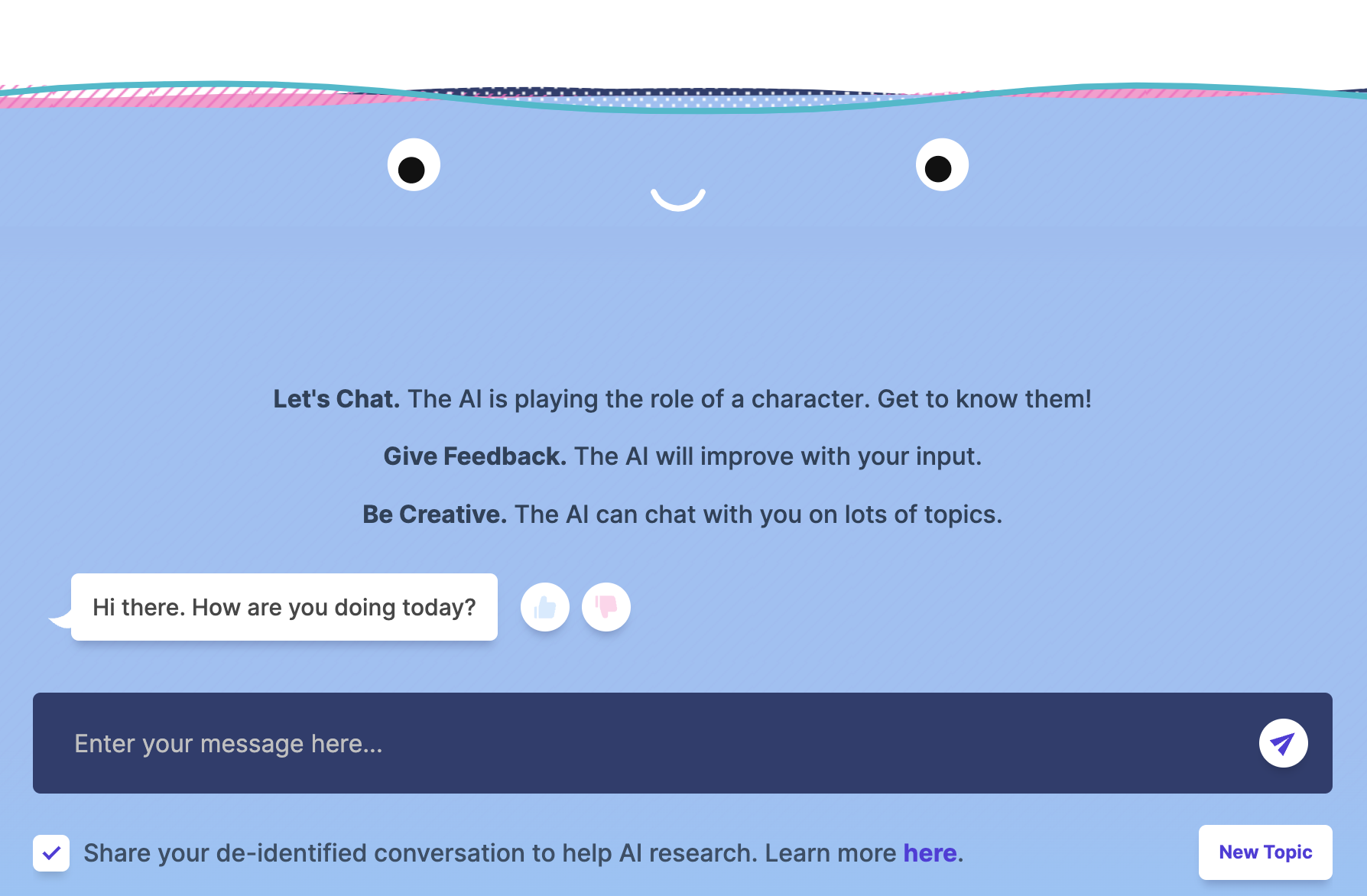}
         \caption{Main and initial chat screen, depicting the abstract face, and extradiegetic \textit{educational} elements aiming to cause the right expectations and mindset for the user (see Sec.~\ref{cogfactors}).}
         \label{fig:1.0}
     \end{subfigure}
     \hfill
     \vspace{1.5cm}
     \begin{subfigure}{0.25\textwidth}
         \centering
         \includegraphics[width=\textwidth]{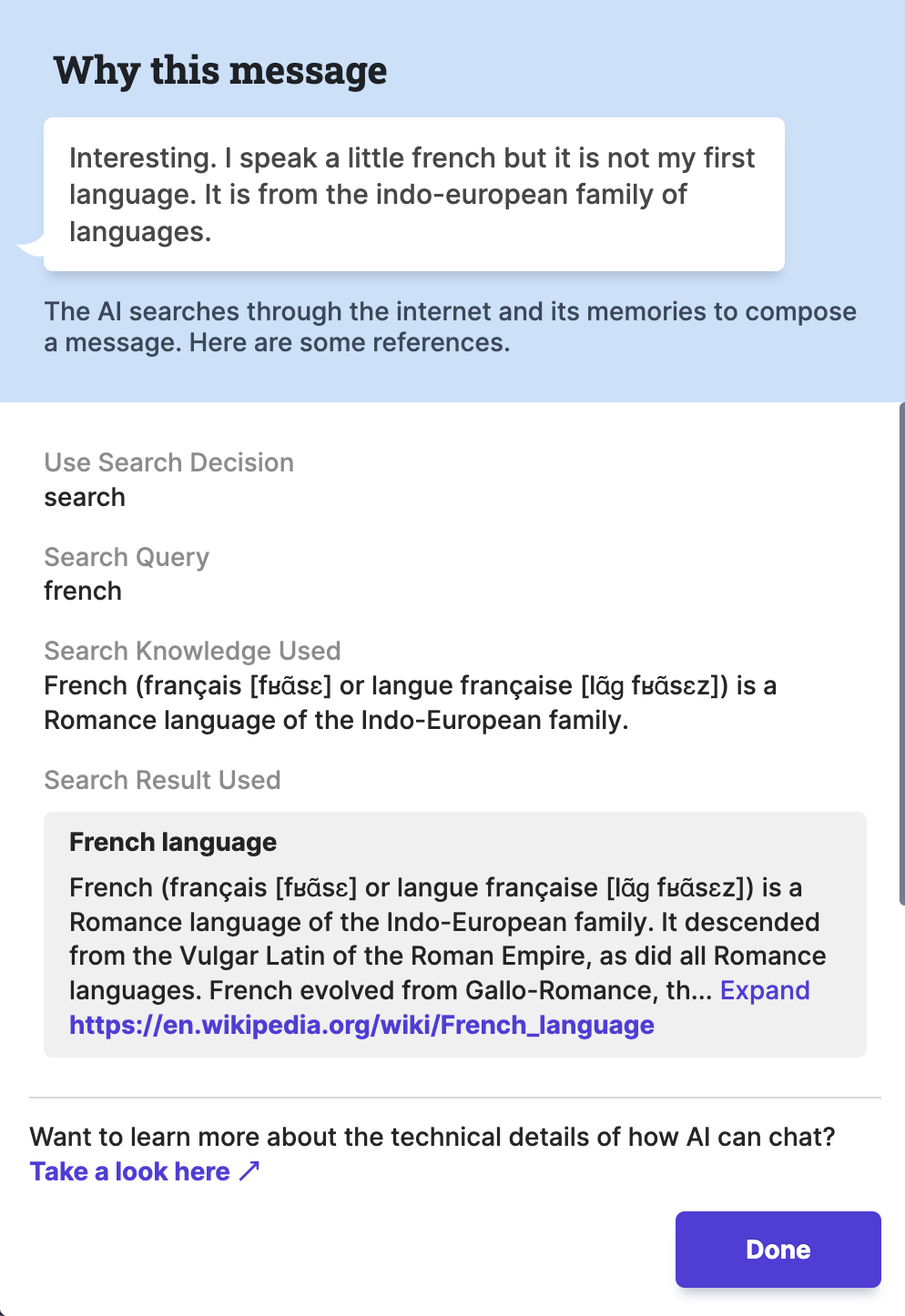}
         \caption{The XAI element ``Why This Message'', shown when the user clicks on an agent's message.}
         \label{fig:1.1}
     \end{subfigure}
     \vspace{1em}%
     \hfill
     \begin{subfigure}{0.2\textwidth}
         \centering
         \includegraphics[width=\textwidth]{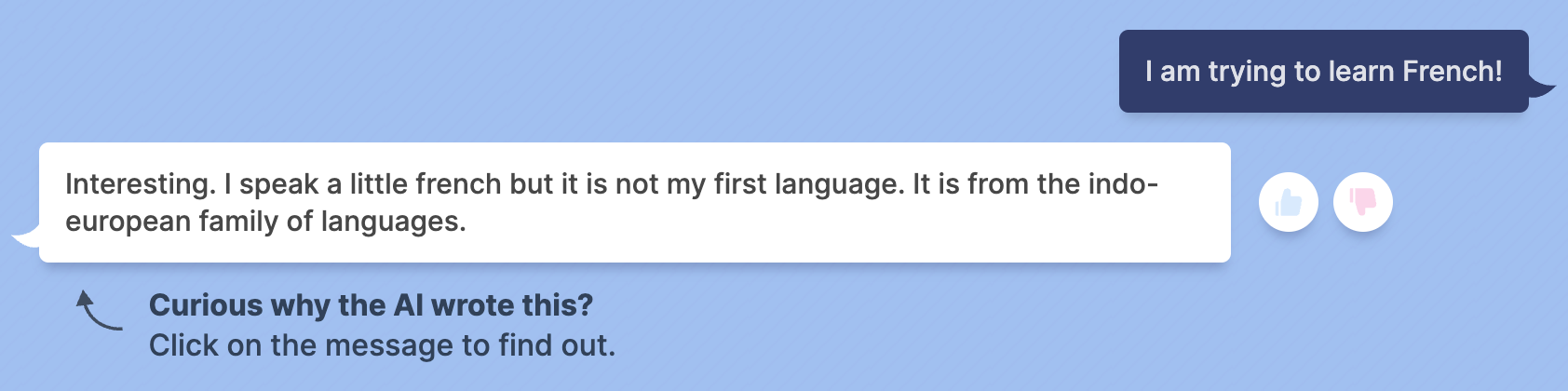}
         \caption{Extradiegetic tone providing \textit{education} to the user about how they can seek explanations about chatbot behavior.}
         \label{fig:1.2}
     \end{subfigure}
     \vspace{1.5cm}
     \hspace{1em}%
     \begin{subfigure}{0.2\textwidth}
         \centering
         \includegraphics[width=\textwidth]{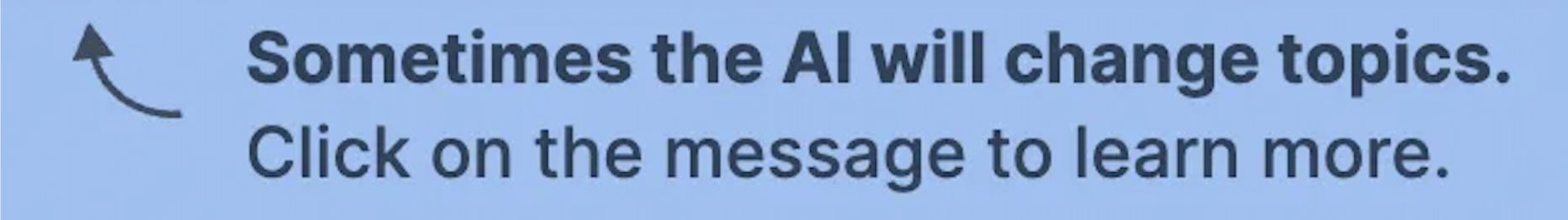}
         \caption{Extradiegetic tone providing \textit{explanation} to the user when a safety classifier is triggered to prevent further chat on an unsafe topic (the user could then click on the previous message to see more context on the safety issue).}
         \label{fig:1.3}
     \end{subfigure}
     \hspace{1em}%
     \vspace{1.5cm}
     \begin{subfigure}{0.2\textwidth}
         \centering
         \includegraphics[width=\textwidth]{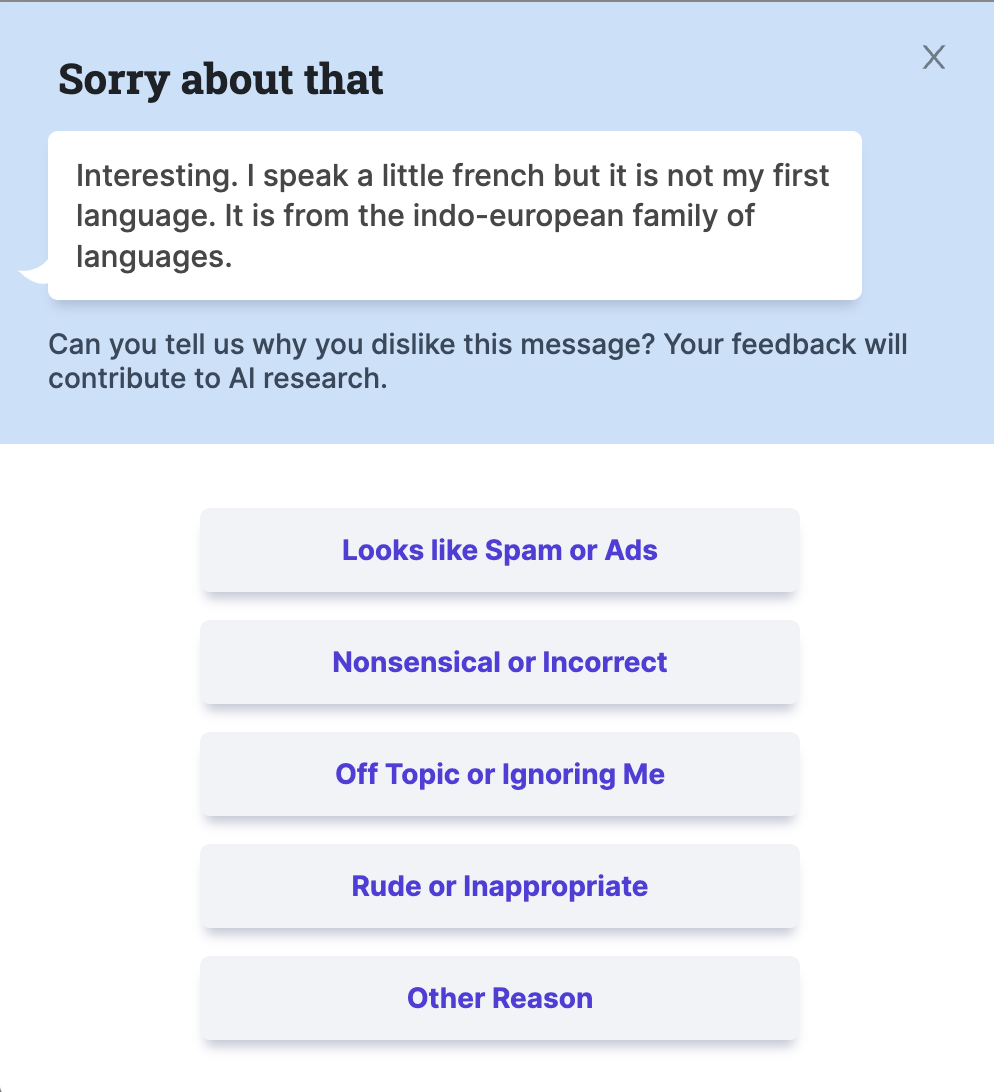}
         \caption{The set of options shown to the user when they click on the thumbs down button for a specific agent message.}
         \label{fig:1.4}
     \end{subfigure}
     \hspace{1em}%
    \begin{subfigure}{0.2\textwidth}
         \centering
         \includegraphics[width=\textwidth]{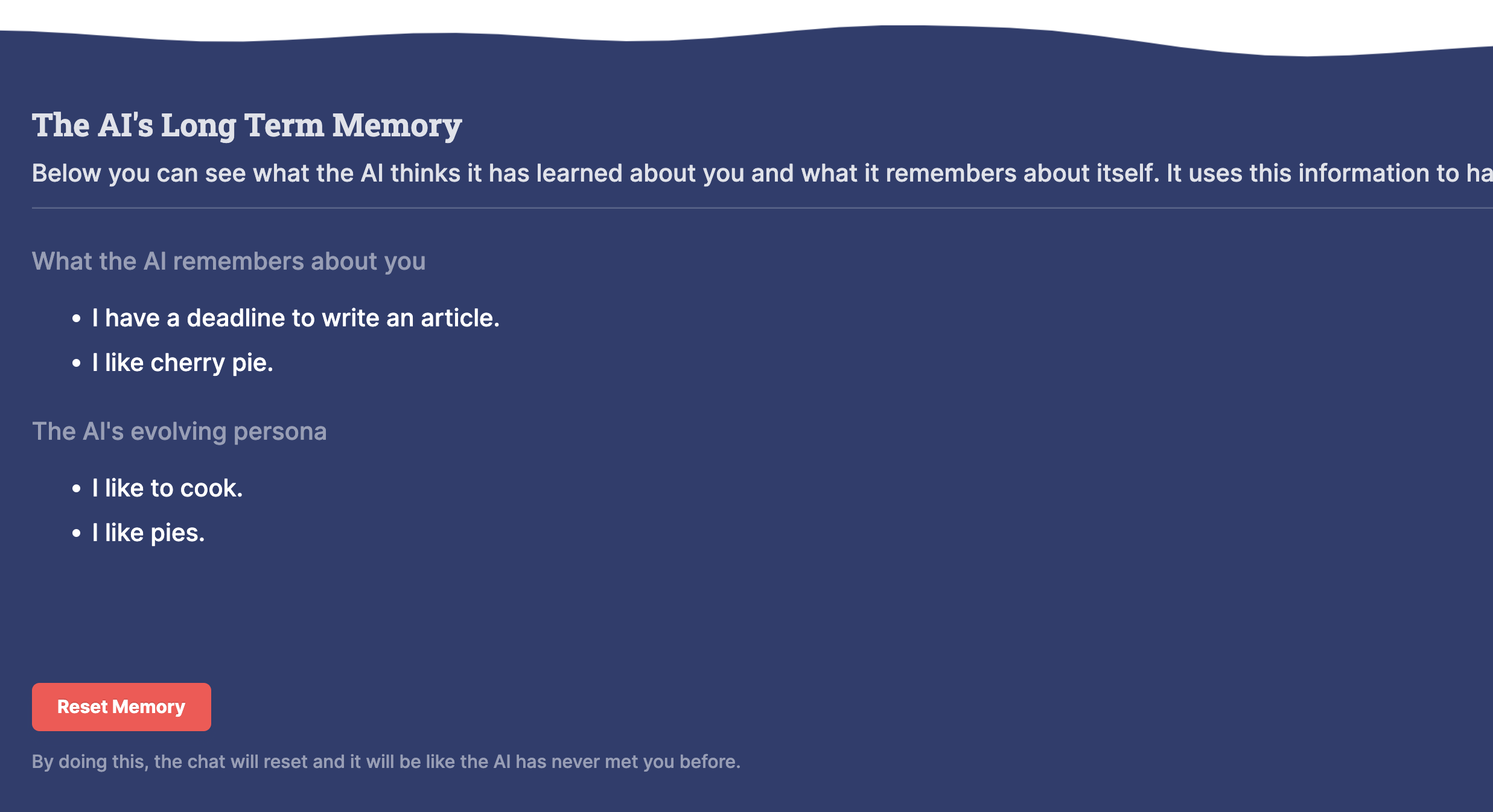}
         \caption{The ``look inside'' page listing the memories the agent has learned or developed about the user or itself, as an attempt to make the system more explainable to the user.}
         \label{fig:1.5}
     \end{subfigure}
     \vspace{-0.5cm}
        \caption{Various screens and interface elements of the chatbot used in the study.}
        \label{fig:chatscreen}
\end{figure}

While we pursued answering the questions as openly as possible in the qualitative format, we also hypothesized a few design solutions to develop a specific interface with which to investigate. The relevant interface elements were: 

\begin{itemize}
    \item \textbf{Why This Message window}: Upon clicking on an agent's message, the user could learn more about how the chatbot had generated a specific message. We hypothesized that such XAI avoidance in a research chatbot may help users learn more about how AI works and boost user's willingness to give feedback and to continue engagement after chatbot failures; see Fig. \ref{fig:1.1}.
    
    \item \textbf{Extradiegetic tone}: Inside the chat, and for various reasons, the system gave the user a message to educate, explain, and contextualize agent's conversational behavior. This explicit extradiegetic tone was also intended as a way to separate user's perceptions of the system and of the agent, to help with the cognitive effects explained in Sec.~\ref{expectationbaseline}; see Figures~\ref{fig:1.2} and \ref{fig:1.3}. 
    
    \item \textbf{An abstract face}: While simplistic and abstract, based on the cognitive factors explained in Sec.~\ref{speakervoid}, we hypothesized that the face seen in Fig.~\ref{fig:1.0} would positively impact user's experience and perceptions of the chat and of the agent, and may also change user behavior in propensity to provide feedback.
\end{itemize}

\subsection{Study Setup and Procedure}

We recruited N=40 participants in the US for their remote participation in 90-minutes sessions. Through a screening survey, we ensured including a diverse set along a few vectors, such as gender (19 female, 21 male), location, familiarity with technology, and ethnicity. In each session, the participant first received a brief introduction to the study, signed a consent form, and was asked to access a URL on the computer. At this stage, participants freely chatted with the chatbot for a minimum of 10 and a maximum of 20 minutes (enforced by the conductors). During this time, participants were asked to think aloud and mention any notable parts of their experience; including about the interface, interaction elements, chatbot messages, urge to give feedback or seek explanation, any points of confusion, etc. After the initial chat, participants were immediately given a 14-questions quantitative questionnaire (5-point Likert scale), asking them about their experience, the interface, their perceptions of the agent and chat, their desire to give feedback or seek explanation, and other topics. In the last and main section of the session, a semi-structured qualitative interview was conducted, based on a pre-written discussion guide, but also including and probing on participant's expressed thoughts during the chats or their answers to the questionnaire. In the end, participants were given a chance to ask any questions they may have, and were compensated 100 USD for their contribution and time.

Embedded in the study, was a small-scale, between-subject experimental setup. Half (N=20) of participants experienced a version of the chatbot that did not include the abstract face shown in Fig. \ref{fig:1.0} while the other half did. The rationale for including an experimental setup exclusive to a hypothesis derived from RQ3 (and not other RQs) was the possibility that the effects of using an abstract face for agent representation would not be experienced consciously enough for the participants to describe them in the think aloud process, questionnaire, or the interview. 

\subsection{Results}

\subsubsection{\textbf{Quantitative results}} All of the questions were mandatory and the response size was N=40. A few of the questions in the questionnaire were directly related to the perceived usefulness of the elements discussed in Sec. \ref{sec:chatbot}, which were hypothesized to address the root cognitive issues (see Sec.~\ref{cogfactors}) and answer the related research questions mentioned earlier. Responses to a question about \textit{the usefulness of the Why This Message (XAI) element} (see Fig.~\ref{fig:1.1}) were  positive ($median=3.5, mean=3.0$ out of 5). Responses to a similar question about the \textit{usefulness of the look inside page} (see Fig.~\ref{fig:1.5}) were more favorable ($median=3.8, mean=4.0$). Responses to another question regarding \textit{participants' willingness to contribute to the AI research through giving feedback} was also positive ($median=3.9, mean=4.0$). Results for \textit{likelihood to seek further conversations with the chatbot} where less positive but favorable ($median=3.1, mean=3.0$). The rest of questionnaire items pertained mostly to the evaluation of the chat experience itself, as a way to expand and orient the quantitative results. As a summary ($medians=$), participants rated the conversation's engagingness ($3.5$), coherence ($3$), and informativeness ($4$); and the chatbot's humanlikeness ($3$), friendliness ($5$), emotional awareness ($3$), intelligence ($3.5$), factual accuracy ($4$), and offensiveness ($1$).

\subsubsection{\textbf{Experiment's results}} On average, participants in the \textit{with face} condition rated the agent more favorably in the questionnaire on every single item mentioned above, except for one. The following table shows the results, where some rows show statistical significance or notable trends.

\begin{table*}[!h]
\caption{Results from the experimental setup (between-subjects, N=20 per condition). Results shown as $mean(median)$. P-values are from a Mann-Whitney U test.}
\label{tab:expres}
\centering
\begin{tabular}{| p{57mm} | c c c |} 
 \hline
 \textit{} &\textbf{ without face} & \textbf{with face} & \textbf{p-value} \\ [0.5ex] 
 \hline
 \textit{usefulness of  Why This Message} & 3.7 (3.5) & 3.5 (3.0) & 0.24 \\
 \hline
 \textit{usefulness of Look Inside} & 3.8 (3.5) & 4 (4.5) & 0.20 \\ 
 \hline
 \textit{willingness for feedback} & 3.6 (4.0) & 4.1 (4.0) & 0.19 \\
 \hline
 \textit{seek more chats} & 2.8 (3.0) & \textbf{3.4 (3.0)} & \textbf{0.09} \\
 \hline
 \textit{engagingness} & 3.2 (3.0) & \textbf{3.8 (4.0)} & \textbf{0.05} \\
 \hline
 \textit{coherence} & 3.0 (3.0) & \textbf{3.6 (4.0)} & \textbf{0.06} \\
 \hline
 \textit{informativeness} & 4.1 (4.0) & \textbf{4.2 (5.0)} & \textbf{0.09} \\
 \hline
 \textit{humanlikeness} & 3.0 (3.0) & \textbf{3.5 (3.0)} & \textbf{0.10} \\
 \hline
 \textit{friendliness} & 4.1 (4.0) & 4.2 (5.0) & 0.32 \\
 \hline
 \textit{emotional awareness} & 2.5 (2.5) & \textbf{3.5 (3.5)} & \textbf{<< 0.01} \\
 \hline
 \textit{intelligence} & 3.2 (3.0) & 3.6 (4.0) & 0.15 \\
 \hline
 \textit{factual accuracy} & 3.6 (4.0) & 3.8 (4.0) & 0.36 \\
 \hline
 \textit{offensiveness} & 1.0 (1.0) & 1.2 (1.0) & 0.28 \\
 \hline
\end{tabular}
\end{table*}

\subsubsection{\textbf{Qualitative results}} Below, some of the main themes of qualitative findings are listed. 

\textbf{Extradiegetic communication. } We found an \textit{extradiegetic} level of message, as a ``voice'' of the system or researchers behind it, and distinct from the \textit{diegetic} messages of the chatbot, enables the interface to clarify, explain, or augment how the chatbot is behaving conversationally. Participants' reactions show that it compartmentalizes their perception. P18 said of the extradiegetic tone: ``\textit{[it's] like a game console; when you [start the experience for the first time], it tells you what some of the controls do. There’s a highlight area [accessible] throughout the chat so you don’t forget}''.
%It also offers education which may drive curiosity and lead to more feedback.

\textbf{UI for conversational control. } Similarly, participants expressed positive sentiment towards an existing ``new topic'' button which allowed for changing topics and reacted positively towards design prototypes used as stimuli, depicting controls that would allow the user to choose specific chat topics from a list, or communicating higher level needs to the system, e.g., feelings or goals. 

% \subsubsection{\textbf{UI for chatbot customization}} slide 11. 

\textbf{XAI usage. } Participants were most interested in seeking an explanation for the chatbot's message when something did not make sense, e.g., a nonsensical response, a mistake, an unnatural transition, or a factual error. 

\textbf{An HCI tension. } While not all participants experienced this to the same extent, seeking XAI elements (e.g., the ``Why This Message'' window) and providing thumbs up/down feedback on specific messages required the participants to interrupt the conversation, cognitively focus on another matter, and redirect their attention. P4 said of the effect on feedback: ``\textit{While I’m having a chat, I’m into it. I’m not looking to [evaluate] it. I won’t [evaluate] it until at least after the chat}''. P39, speaking about the XAI elements, said: ``\textit{I’m just paying attention to the conversation right now.}''. A few design solutions are discussed in the next section.

\textbf{Opinions about the face. } Interestingly, and as opposed to the quantitative results, most participants did not consciously perceive the abstract face to affect their experience. P36 said: ``\textit{I didn’t really notice it, but I like it.}'' 

\textbf{The word ``bot''. } Some participants described confusion around what the ``chatbot'' is capable of, likely compounding the root cognitive causes explained in Sec.~\ref{cogfactors}. The word ``chatbot'' has a customer service and strongly utilitarian connotation in English. This is especially an issue if the chatbot is aimed at being sociable. P23 said of ideal naming: ``\textit{A name like ‘Buddy’ or ‘Friend,’ that shows how this is not just an agent to do a service}''.

\section{Discussion and Design Recommendations}

Overall, the study results confirms that the interaction and UI design for research chatbots is both non-trivial and consequential, and we observed user confusions that pertain to the cognitive factors previously outlined (see Sec. \ref{cogfactors}). Intentionally designing the interface, and adding specific features to address the cognitive factors can positively impact user perception, user behavior, and the resulting data generated. A few design directions and specific recommendations derived from our study include: avoiding the term ``bot'', including an extradiegetic level of communication about chatbot behavior and system actions, allowing for communication control through non-verbal elements, including subtle anthropomorphism (especially for sociable chatbots), and using XAI.

As previously mentioned, offering XAI and even expecting user feedback (e.g. thumbs up/down) can result in a tension, where rather ironically, the more engaging and captivating the conversation is, the less likely the user would be to break its flow and take a secondary action. This issue implies a design challenge. Interestingly, and while it was not statistically significant (p=0.2), we observed a \textit{reduced} expressed willingness to view the ``Why This Message'' XAI element in the experimental condition \textit{with a face} (see Table~\ref{tab:expres}). Given that the presence of the face showed a significant increase in perceived engagingness, this may be a trend showing how a more engaged conversation \textit{reduces} the likelihood for seeking XAI or providing feedback. Possible design solutions for the above issue include proactive explanations (e.g., in the case of model failure) or surfacing such UI elements in between topic changes or when the user seems less actively engaged. It would be interesting to experiment with diegetic conversational interfaces for XAI or feedback entry (e.g., ``why did you say that?'' or ``how would you rate my precious response?''), especially if the chatbot does not have a persona or a character facade.

Lastly, we observed that triangulation of the quantitative findings (e.g., qualitatively expanding on perceptions of ``engagingness'' or ``humanlikeness'') can offer specific and actionable insights that should accompany large-scale experimental setups asking crowd workers the same quantitatively. With increased quality of generations from recent large models, abstract quantitative questions may become subjective, and ultimately less converging \cite{chatbotEvalOpenProblem}. While this was not a focus in our work, the ``orientations'' in \textit{why} a specific model is perceived as more or less engaging was enlightening in our work, because it highlighted reasons why engagement levels changes; e.g., transitions between topics was not natural, or that the user had a misconception about what the chatbot is capable of. 

\section{Limitations}

Many of the experimental results do not reach statistical significance but do show trends. A higher number of participants may further clarify results. Moreover, using additional conditions with variations on the agent representation may offer more design guidelines. Furthermore, user perceptions are likely to change in long-term use of the chatbot, compared to one-time interactions. We aim to focus on long-term relationship in subsequent works \cite{skjuve2022longitudinal, replika}.

\section{Conclusions}

We propose a focus on the HCI elements of the public deployment of research chatbots. We argue that design decisions involved are non-trivial and affect both user's experience and the deployment goals (e.g., feedback rates). We lay out a grounding in cognitive science and report on a mixed-methods user study to answer some of the research questions posed. We propose a two-layered \textit{diegetic} and \textit{extradiegetic} communication. We find that an abstract anthropomorphic visual representation of the chatbot to positively affect user perception. Lastly, we find XAI to be helpful, but require innovative design solutions to be compatible with the sequential nature of the conversation.

\bibliographystyle{acl_natbib}
\bibliography{sample-base}

\end{document}